\documentclass[sigconf]{acmart}

\usepackage[font={small}]{caption}

\renewcommand\footnotetextcopyrightpermission[1]{} 
\AtBeginDocument{%
  \providecommand\BibTeX{{%
    \normalfont B\kern-0.5em{\scshape i\kern-0.25em b}\kern-0.8em\TeX}}}

\begin{document}

\title{De-identification of Unstructured Clinical Texts from Sequence to Sequence Perspective}

\author{Md. Monowar Anjum}
\email{anjumm1@myumanitoba.ca}
\affiliation{%
  \institution{Department of Computer Science\\University of Manitoba}
  \city{Winnipeg}
  \state{MB}
  \country{Canada}
}

\author{Noman Mohammed}
\email{noman@cs.umanitoba.ca}
\affiliation{%
  \institution{Department of Computer Science\\University of Manitoba}
  \city{Winnipeg}
  \state{MB}
  \country{Canada}
}

\author{Xiaoqian Jiang}
\email{xiaoqian.jiang@uth.tmc.edu}
\affiliation{%
  \institution{School of Biomedical Informatics\\University of Texas}
  \city{Houston}
  \state{TX}
  \country{USA}
}

% \renewcommand{\shortauthors}{Anjum et al.}

\begin{abstract}
In this work, we propose a novel problem formulation for de-identification of unstructured clinical text. We formulate the de-identification problem as a sequence to sequence learning problem instead of a token classification problem. Our approach is inspired by the recent state-of -the-art performance of sequence to sequence learning models for named entity recognition. Early experimentation of our proposed approach achieved 98.91\% recall rate on i2b2 dataset. This performance is comparable to current state-of-the-art models for unstructured clinical text de-identification. 
\end{abstract}





\maketitle

\section{Introduction}

A recent study~\cite{toscano2018electronic} found out that about 95\% of the eligible hospitals and 62\% of all office-based physicians used Electronic Health Records (EHR) systems. The EHR records contain patient information, history, medication prescription, vital signs, laboratory test results etc. Moreover, they contain unstructured notes from physicians regarding the patient which are rich sources of contextual medical information. Statistical analysis of EHR can lead to less error-prone medical diagnosis, healthcare cost reduction and better short term preventive care~\cite{fernandez2013security}. 

However, the analysis of EHR for various clinically significant statistical task is not straightforward. EHRs often contain sensitive identifying information regarding patients which are generally considered private. To be specific, the U.S. Health Insurance Portability and Accountability Act (HIPAA) requires 18 re-identifying categories of information to be sanitized from EHR before dissemination~\cite{hipaa2021}. These categories are called personal health information (PHI) and include patients name, profession, unique identifying numbers such as social security number, driving license number or medical insurance number etc. HIPAA  precludes researchers from performing analysis on large scale EHR repositories unless they are de-identified.

Using human annotators for de-identifying EHR is a costly process. One estimate by~\cite{douglass2005identification} reported a cost of \$50/hour for human annotators who read around 20000 words per hour. At this rate, it will cost \$250,000 to annotate a dataset that have 100 million words. Moreover, this EHR de-identification by human annotation is also error prone. Prior work~\cite{neamatullah2008automated} reported that recall value varied between 63-94\% for 13 human annotators when they were asked to de-identify approximately 130 patient notes. Therefore,  it is likely that multiple human annotators will be required to annotate the same patient notes to ensure the quality of de-identified text. In that case, the cost of de-identification will increase even more. 

The shortcomings of manual EHR de-identification process led to extensive research in the domain of automated EHR de-identification model. Before the widespread adoption of deep learning in natural language processing (NLP) tasks, majority of automated EHR de-identification systems adopted rule-based approach~\cite{douglass2005identification}. Rule-based approaches suffer from multiple shortcomings. Firstly, they are sensitive to dataset. Rules that work for one dataset may require extensive calibration for a different dataset. Secondly, rules fail to take context into account. Consequently, rule based approaches are not scalable and have poor performance generally. Supervised machine learning methods offered comparatively better performance and generalizability compared to rule-based methods. These models solve the de-identification problem as binary (PHI and non-PHI) or categorical (types of PHI) classification problem. The shortcoming of these systems is that they rely on handcrafted features which are extracted from the tokens or sentences. Deep learning solves this dependency by learning useful features from text data without any human intervention. Deep learning methods that use non-linear neural network models have shown state-of-the-art performance in natural language understanding tasks such as named entity recognition~\cite{lample-etal-2016-neural}, part of speech tagging~\cite{gui2017part}, neural machine translation~\cite{bahdanau2014neural} etc. All of these tasks have been solved by some form of recurrent neural network (RNN). In 2018, Google proposed BERT~\cite{devlin-etal-2019-bert}, a state-of-the art neural network architecture that set the benchmark for a range of NLP tasks. BERT used a new mechanism called ``self-attention''. The current state-of-the-art EHR de-identification model~\cite{ahmed2020identification} also employs attention based approach to get improved results in benchmark datasets.\\

\noindent \textbf{Motivation.} During literature review, we noted that the earliest EHR de-identification model is dated back in 1996 which was called SCRUB~\cite{sweeney1996replacing}. Subsequent 25 years of academic research produced a large body of work for EHR de-identification. During this period, methods gradually progressed from rule based to deep learning models. Proliferation of benchmark datasets such as i2b2~\cite{stubbs2015automated}, MIMIC-II~\cite{lee2011open} and MIMIC-III~\cite{johnson2016mimic} also played a pivotal role in this development. Current state-of-the-art~\cite{ahmed2020identification} achieves a recall rate (de-identification accuracy) of 98.41\%, 82.9\% and 100.0\% respectively for these datasets. We observe that recall rate for the MIMIC-II dataset is particularly low. Therefore, we believe there is room for improvement in terms of model performance on benchmark datasets.
% The result can be interpreted as 13\% of the clinical notes will still contain personal health information tokens in them even after they are sanitized by the model. 
Moreover, prior works modeled clinical text de-identification as a classification problem. \emph{We note that de-identification problem can be modeled differently as well (i.e., sequence to sequence modeling, metric learning). However, these approaches have not been explored in literature yet.}\\
\begin{figure}[!t]
    \centering
    \includegraphics[width=0.45\textwidth]{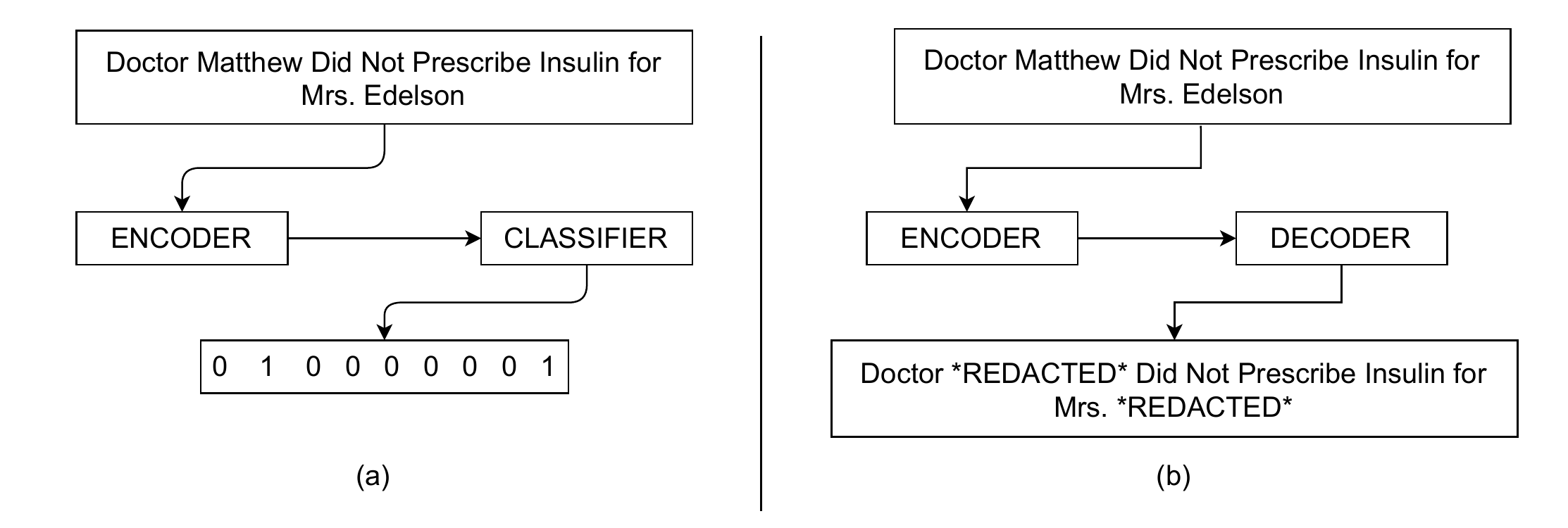}
    \caption{Classification vs. Sequence to Sequence Modeling}
    \label{fig:comparison_of_solution}
\end{figure}
\noindent \textbf{Contribution.} We model the de-identification problem as a sequence to sequence learning problem. Our approach is inspired by the recent advancement in named entity recognition in~\cite{chen2018learning} which combined transfer learning and sequence to sequence modeling. To the best of our knowledge, this is the first work that models de-identification problem as a sequence to sequence learning problem. The summary of our contributions are:
\begin{itemize}
    \item We propose a novel sequence-to-sequence learning based formulation of the unstructured clinical text de-identification problem. 
    \item Our proposed method achieves 98.91\% recall for i2b2 dataset.
\end{itemize}
\section{Problem Definition}
The problem we tackle in this work is the unstructured clinical text de-identification problem. Let us consider a unstructured clinical text $T$ which consists of $k$ tokens $\{t_1,t_2,\cdots,t_k\}$. HIPAA identifies 18 types of PHI which are needed to be removed from EHR before public release. Each token $t_i\in T$ either belongs to one of the classes of PHI or not. 

Consider a function $\mathcal{F}$. It takes $T$ as input and produces a sequence of $k$ tokens, $\hat{T}=\{\hat{t_1},\hat{t_2},\cdots,\hat{t_k}\}$. Equation \ref{eq:definition} defines each $\hat{t_i}\in \hat{T}$.
\begin{equation}
    \hat{t_i} = \begin{cases}
    t_i , & \text{if } t_i \notin \text{ PHI}\\
    t_{\text{special}} , & \text{if } t_i \in \text{ PHI}
    \end{cases}
    \label{eq:definition}
\end{equation}
$t_{\text{special}}$ is a special token to replace the PHI tokens. It can be same for all PHI classes or different for each PHI classes. The objective of the de-identification problem is to find the optimal $\mathcal{F}$ so that $\mathcal{F}(T)=\hat{T}$ results in maximum precision and recall.   
\section{Proposed Solution}

We propose a encoder-decoder architecture for unstructured clinical text de-identification. This is the standard architecture in multiple natural language processing tasks such as named entity recognition~\cite{chen2018learning}, machine translation~\cite{bahdanau2014neural}, etc. Both encoder and decoder consist of ``multi head self-attention'' layers. From the encoder perspective, the attention layers encode a variable length input sequence into a fixed length context vector. Similarly, from the decoder perspective, the attention layers decode a fixed length context vector to a variable length output sequence. During the training process, the model learns about the context and features of the PHI tokens. The decoder then maps the non-PHI tokens to their input and the PHI tokens to the special token.  

Figure 1(a) shows the problem formulations used in prior works. The de-identification model takes a sequence of tokens as input and output is a vector that contains the label for each token. 0 stands for non-PHI and 1 stands for PHI. On the other hand, Figure 1(b) shows our proposed solution model. The input is the same as prior works. The output is a sequence of tokens where the input PHI-tokens have been replaced with a special token \verb|REDACTED|. This is a fundamental difference in our approach compared to the prior works. In prior approaches, the tokens were mapped to only two classes. In our approach, each token is mapped to \emph{itself} if it is a non-PHI token. If it is a PHI token, then it is mapped to a special token. In other words, the output set of prior approaches are \{0,1\}. In our approach, the output set consists of all possible input tokens and the special tokens.

An important difference between our approach and the prior approaches is the model architecture. The current deep learning models for de-identification learn a task specific encoding before applying that encoding for token classification~\cite{ahmed2020identification, dernoncourt2017identification}. Therefore, current architectures are encoder-classifier models. On the contrary, our proposed architecture is an encoder-decoder model. The encoder model learns to encode unstructured clinical texts. The decoder model learns to replace the PHI tokens with special token. In other words, the decoder model \emph{\textbf{translates}} the input sequence to a de-identified output sequence. 

Let us explain this difference with an example. In Figure 1(a), the non PHI tokens \{\verb|Doctor, Did, Not, Prescribe, Insulin, For,| \verb|Mrs|\} are mapped to 0 and PHI tokens \{\verb|Matthew, Edelson|\} are mapped to 1. Let's consider the example in Figure 1(b). We assume that each token in the input sequence is mapped to id 1-9 and the token \verb|REDACTED| is mapped to id 10. In the output sequence, each non-PHI tokens are mapped to their own token ids while the PHI tokens are mapped to id 10. In other words, the input sequence ``Doctor Matthew did not prescribe insulin for Mrs. Edelson'' becomes \{1,2,3,4,5,6,7,8,9\} before being fed into encoder. The decoder produces \{1,10,3,4,5,6,7,8,10\} which is mapped back to ``Doctor REDACTED did not prescribe insulin for Mrs. REDACTED''.
\section{Preliminary Result}
\textbf{Implementation.} We implemented the encoder-decoder model for sequence to sequence learning in python. We used Tensorflow 2.0 as implementation framework. Both encoder and decoder model consists of an embedding layer and 8 multi-head self-attention layers. We used ``weighted softmax crossentropy'' as loss function. ADAM optimizer was used for training optimization. We set the initial learning rate to 0.002. 

We evaluate the performance of our proposed model based on recall value. Recall value represents how many of the PHI tokens have been replaced by the special tokens in the output sequences. We put higher emphasis on recall than precision\footnote{This is because in the worst case scenario, a single PHI token missed by the model can re-identify the clinical text}.\\

\noindent \textbf{Experimental Result.} We perform evaluation of our proposed model on the i2b2 2014 challenge task dataset. We present our result in Table \ref{tab:result}. We achieved higher recall and F1-score than the current state-of-the-art (SOTA) model in literature. However, our precision score is slightly less than SOTA. We note that, the number of parameters in our model is comparatively lower than SOTA which may have played a role in this case. We aim to increase number of model parameters to achieve better precision in future iterations. We are currently performing experimentation on MIMIC-II and MIMIC-III datasets. The results from these experiments will be included in our future work.  

\section{Future Work}
\textbf{Experiment on MIMIC-II and III.} We will perform experiment on MIMIC-II and III datasets. During literature review, we noted that current models have very low recall rate in MIMIC-II. We aim to improve that recall rate. Moreover, we plan to generate additional unstructured text data from MIMIC-II and III dataset and train our model on it to achieve greater robustness.

\noindent \textbf{Transfer Learning.} We note that named entity recognition problem has similarity with unstructured clinical text de-identification. Literature review revealed that transfer learning improves the accuracy and generalizability of named entity recognition models. To the best of our knowledge transfer learning has not been leveraged for clinical text de-identification problem. We intend to explore this direction in our future works.

\noindent \textbf{Semi Supervised Learning.} An important aspect of de- identification model training is the unavailability of ground truth data. Manual ground truth data annotation is an expensive and error-prone process. Semi-supervised learning can help to mitigate this challenge by training models on partially labeled dataset. Literature review of de-identification models reveal that semi-supervised learning have not been leveraged yet for clinical text de-identification. We will explore this training method for our models in our future works.

\noindent \textbf{Domain Adaptation.} A common problem among current de- identification models is that domain shift issue. Models trained on one dataset generally perform very poorly on other datasets. Tzeng et al. showed that this issue can be mitigated by training in the models using adversarial domain adaptation~\cite{tzeng2017adversarial}. We intend to explore this training approach in future iterations of this work. 

\section{Conclusion}
In this work we presented a novel sequence-to-sequence problem formulation for the clinical text de-identification problem. We designed an encoder-decoder architecture to translated unstructured clinical texts with PHI tokens to sanitized clinical texts without the PHI tokens. Preliminary analysis of our proposed method shows promising evaluation metric scores. We are currently experimenting on other benchmark datasets to assess the effectiveness of our problem formulation and solution model.   
\begin{table}[t]
    \centering
    \scriptsize
    \caption{Preliminary experimental evaluation of our proposed method on i2b2 2014 dataset. The best results are marked in bold.}
    \begin{tabular}{|c|c|c|c|c|}
    \hline
        Method &\# parameters & Precision & Recall & F1-Score \\
        \hline
        \hline
        Dernoncourt et al.~\cite{dernoncourt2017identification} & N/A & 97.92 & 97.84 & 97.88 \\
        \hline
        Khin et al.~\cite{khin2018deep} & N/A & 98.30 & 97.37 & 97.83 \\
        \hline
        Tanbir et al.~\cite{ahmed2020identification} (SOTA) & 110,000,000 &\textbf{99.01} & 98.41 & 98.22\\ 
        \hline
        Proposed Method & 78,000,000 & 98.12 & \textbf{98.91} & \textbf{98.51} \\ 
        \hline
    \end{tabular}
    \label{tab:result}
\end{table}

\bibliographystyle{ACM-Reference-Format}
\bibliography{sample-base}


\begin{thebibliography}{18}

%%% the Web address.

\ifx \showCODEN    \undefined \def \showCODEN     #1{\unskip}     \fi
\ifx \showDOI      \undefined \def \showDOI       #1{#1}\fi
\ifx \showISBNx    \undefined \def \showISBNx     #1{\unskip}     \fi
\ifx \showISBNxiii \undefined \def \showISBNxiii  #1{\unskip}     \fi
\ifx \showISSN     \undefined \def \showISSN      #1{\unskip}     \fi
\ifx \showLCCN     \undefined \def \showLCCN      #1{\unskip}     \fi
\ifx \shownote     \undefined \def \shownote      #1{#1}          \fi
\ifx \showarticletitle \undefined \def \showarticletitle #1{#1}   \fi
\ifx \showURL      \undefined \def \showURL       {\relax}        \fi
% invisible to TeX
\providecommand\bibfield[2]{#2}
\providecommand\bibinfo[2]{#2}
\providecommand\natexlab[1]{#1}
\providecommand\showeprint[2][]{arXiv:#2}

\bibitem[\protect\citeauthoryear{Ahmed, Al~Aziz, and Mohammed}{Ahmed
  et~al\mbox{.}}{2020}]%
        {ahmed2020identification}
\bibfield{author}{\bibinfo{person}{Tanbir Ahmed}, \bibinfo{person}{Md~Momin
  Al~Aziz}, {and} \bibinfo{person}{Noman Mohammed}.}
  \bibinfo{year}{2020}\natexlab{}.
\newblock \showarticletitle{De-identification of electronic health record using
  neural network}.
\newblock \bibinfo{journal}{\emph{Scientific reports}} \bibinfo{volume}{10},
  \bibinfo{number}{1} (\bibinfo{year}{2020}), \bibinfo{pages}{1--11}.
\newblock


\bibitem[\protect\citeauthoryear{Bahdanau, Cho, and Bengio}{Bahdanau
  et~al\mbox{.}}{2015}]%
        {bahdanau2014neural}
\bibfield{author}{\bibinfo{person}{Dzmitry Bahdanau},
  \bibinfo{person}{Kyunghyun Cho}, {and} \bibinfo{person}{Yoshua Bengio}.}
  \bibinfo{year}{2015}\natexlab{}.
\newblock \showarticletitle{Neural machine translation by jointly learning to
  align and translate}. In \bibinfo{booktitle}{\emph{Proccedings of
  International Conference On Learning Representations}}.
\newblock


\bibitem[\protect\citeauthoryear{Chen and Moschitti}{Chen and
  Moschitti}{2018}]%
        {chen2018learning}
\bibfield{author}{\bibinfo{person}{Lingzhen Chen} {and}
  \bibinfo{person}{Alessandro Moschitti}.} \bibinfo{year}{2018}\natexlab{}.
\newblock \showarticletitle{Learning to progressively recognize new named
  entities with sequence to sequence models}. In
  \bibinfo{booktitle}{\emph{Proceedings of the 27th International Conference on
  Computational Linguistics}}. \bibinfo{pages}{2181--2191}.
\newblock


\bibitem[\protect\citeauthoryear{Dernoncourt, Lee, Uzuner, and
  Szolovits}{Dernoncourt et~al\mbox{.}}{2017}]%
        {dernoncourt2017identification}
\bibfield{author}{\bibinfo{person}{Franck Dernoncourt},
  \bibinfo{person}{Ji~Young Lee}, \bibinfo{person}{Ozlem Uzuner}, {and}
  \bibinfo{person}{Peter Szolovits}.} \bibinfo{year}{2017}\natexlab{}.
\newblock \showarticletitle{De-identification of patient notes with recurrent
  neural networks}.
\newblock \bibinfo{journal}{\emph{Journal of the American Medical Informatics
  Association}} \bibinfo{volume}{24}, \bibinfo{number}{3}
  (\bibinfo{year}{2017}), \bibinfo{pages}{596--606}.
\newblock


\bibitem[\protect\citeauthoryear{Devlin, Chang, Lee, and Toutanova}{Devlin
  et~al\mbox{.}}{2019}]%
        {devlin-etal-2019-bert}
\bibfield{author}{\bibinfo{person}{Jacob Devlin}, \bibinfo{person}{Ming-Wei
  Chang}, \bibinfo{person}{Kenton Lee}, {and} \bibinfo{person}{Kristina
  Toutanova}.} \bibinfo{year}{2019}\natexlab{}.
\newblock \showarticletitle{{BERT}: Pre-training of Deep Bidirectional
  Transformers for Language Understanding}. In
  \bibinfo{booktitle}{\emph{Proceedings of the 2019 Conference of the North
  {A}merican Chapter of the Association for Computational Linguistics: Human
  Language Technologies, Volume 1 (Long and Short Papers)}}.
  \bibinfo{publisher}{Association for Computational Linguistics},
  \bibinfo{address}{Minneapolis, Minnesota}, \bibinfo{pages}{4171--4186}.
\newblock
\urldef\tempurl%
\url{https://doi.org/10.18653/v1/N19-1423}
\showDOI{\tempurl}


\bibitem[\protect\citeauthoryear{Douglass, Cliffford, Reisner, Long, Moody, and
  Mark}{Douglass et~al\mbox{.}}{2005}]%
        {douglass2005identification}
\bibfield{author}{\bibinfo{person}{MM Douglass}, \bibinfo{person}{GD
  Cliffford}, \bibinfo{person}{Andrew Reisner}, \bibinfo{person}{WJ Long},
  \bibinfo{person}{GB Moody}, {and} \bibinfo{person}{RG Mark}.}
  \bibinfo{year}{2005}\natexlab{}.
\newblock \showarticletitle{De-identification algorithm for free-text nursing
  notes}. In \bibinfo{booktitle}{\emph{Computers in Cardiology, 2005}}. IEEE,
  \bibinfo{pages}{331--334}.
\newblock


\bibitem[\protect\citeauthoryear{Fern{\'a}ndez-Alem{\'a}n, Se{\~n}or, Lozoya,
  and Toval}{Fern{\'a}ndez-Alem{\'a}n et~al\mbox{.}}{2013}]%
        {fernandez2013security}
\bibfield{author}{\bibinfo{person}{Jos{\'e}~Luis Fern{\'a}ndez-Alem{\'a}n},
  \bibinfo{person}{Inmaculada~Carri{\'o}n Se{\~n}or}, \bibinfo{person}{Pedro
  {\'A}ngel~Oliver Lozoya}, {and} \bibinfo{person}{Ambrosio Toval}.}
  \bibinfo{year}{2013}\natexlab{}.
\newblock \showarticletitle{Security and privacy in electronic health records:
  A systematic literature review}.
\newblock \bibinfo{journal}{\emph{Journal of biomedical informatics}}
  \bibinfo{volume}{46} (\bibinfo{year}{2013}), \bibinfo{pages}{541--562}.
\newblock


\bibitem[\protect\citeauthoryear{Gui, Zhang, Huang, Peng, and Huang}{Gui
  et~al\mbox{.}}{2017}]%
        {gui2017part}
\bibfield{author}{\bibinfo{person}{Tao Gui}, \bibinfo{person}{Qi Zhang},
  \bibinfo{person}{Haoran Huang}, \bibinfo{person}{Minlong Peng}, {and}
  \bibinfo{person}{Xuan-Jing Huang}.} \bibinfo{year}{2017}\natexlab{}.
\newblock \showarticletitle{Part-of-speech tagging for twitter with adversarial
  neural networks}. In \bibinfo{booktitle}{\emph{Proceedings of the 2017
  Conference on Empirical Methods in Natural Language Processing}}.
  \bibinfo{pages}{2411--2420}.
\newblock


\bibitem[\protect\citeauthoryear{Johnson, Pollard, Shen, Li-Wei, Feng,
  Ghassemi, Moody, Szolovits, Celi, and Mark}{Johnson et~al\mbox{.}}{2016}]%
        {johnson2016mimic}
\bibfield{author}{\bibinfo{person}{Alistair~EW Johnson}, \bibinfo{person}{Tom~J
  Pollard}, \bibinfo{person}{Lu Shen}, \bibinfo{person}{H~Lehman Li-Wei},
  \bibinfo{person}{Mengling Feng}, \bibinfo{person}{Mohammad Ghassemi},
  \bibinfo{person}{Benjamin Moody}, \bibinfo{person}{Peter Szolovits},
  \bibinfo{person}{Leo~Anthony Celi}, {and} \bibinfo{person}{Roger~G Mark}.}
  \bibinfo{year}{2016}\natexlab{}.
\newblock \showarticletitle{MIMIC-III, a freely accessible critical care
  database}.
\newblock \bibinfo{journal}{\emph{Scientific data}} \bibinfo{volume}{3},
  \bibinfo{number}{1} (\bibinfo{year}{2016}), \bibinfo{pages}{1--9}.
\newblock


\bibitem[\protect\citeauthoryear{Khin, Burckhardt, and Padman}{Khin
  et~al\mbox{.}}{2018}]%
        {khin2018deep}
\bibfield{author}{\bibinfo{person}{Kaung Khin}, \bibinfo{person}{Philipp
  Burckhardt}, {and} \bibinfo{person}{Rema Padman}.}
  \bibinfo{year}{2018}\natexlab{}.
\newblock \showarticletitle{A deep learning architecture for de-identification
  of patient notes: Implementation and evaluation}.
\newblock \bibinfo{journal}{\emph{arXiv preprint arXiv:1810.01570}}
  (\bibinfo{year}{2018}).
\newblock


\bibitem[\protect\citeauthoryear{Lample, Ballesteros, Subramanian, Kawakami,
  and Dyer}{Lample et~al\mbox{.}}{2016}]%
        {lample-etal-2016-neural}
\bibfield{author}{\bibinfo{person}{Guillaume Lample}, \bibinfo{person}{Miguel
  Ballesteros}, \bibinfo{person}{Sandeep Subramanian}, \bibinfo{person}{Kazuya
  Kawakami}, {and} \bibinfo{person}{Chris Dyer}.}
  \bibinfo{year}{2016}\natexlab{}.
\newblock \showarticletitle{Neural Architectures for Named Entity Recognition}.
  In \bibinfo{booktitle}{\emph{Proceedings of the 2016 Conference of the North
  {A}merican Chapter of the Association for Computational Linguistics: Human
  Language Technologies}}. \bibinfo{publisher}{Association for Computational
  Linguistics}, \bibinfo{address}{San Diego, California},
  \bibinfo{pages}{260--270}.
\newblock
\urldef\tempurl%
\url{https://doi.org/10.18653/v1/N16-1030}
\showDOI{\tempurl}


\bibitem[\protect\citeauthoryear{Lee, Scott, Villarroel, Clifford, Saeed, and
  Mark}{Lee et~al\mbox{.}}{2011}]%
        {lee2011open}
\bibfield{author}{\bibinfo{person}{Joon Lee}, \bibinfo{person}{Daniel~J Scott},
  \bibinfo{person}{Mauricio Villarroel}, \bibinfo{person}{Gari~D Clifford},
  \bibinfo{person}{Mohammed Saeed}, {and} \bibinfo{person}{Roger~G Mark}.}
  \bibinfo{year}{2011}\natexlab{}.
\newblock \showarticletitle{Open-access MIMIC-II database for intensive care
  research}. In \bibinfo{booktitle}{\emph{2011 Annual International Conference
  of the IEEE Engineering in Medicine and Biology Society}}. IEEE,
  \bibinfo{pages}{8315--8318}.
\newblock


\bibitem[\protect\citeauthoryear{Neamatullah, Douglass, Li-wei, Reisner,
  Villarroel, Long, Szolovits, Moody, Mark, and Clifford}{Neamatullah
  et~al\mbox{.}}{2008}]%
        {neamatullah2008automated}
\bibfield{author}{\bibinfo{person}{Ishna Neamatullah},
  \bibinfo{person}{Margaret~M Douglass}, \bibinfo{person}{H~Lehman Li-wei},
  \bibinfo{person}{Andrew Reisner}, \bibinfo{person}{Mauricio Villarroel},
  \bibinfo{person}{William~J Long}, \bibinfo{person}{Peter Szolovits},
  \bibinfo{person}{George~B Moody}, \bibinfo{person}{Roger~G Mark}, {and}
  \bibinfo{person}{Gari~D Clifford}.} \bibinfo{year}{2008}\natexlab{}.
\newblock \showarticletitle{Automated de-identification of free-text medical
  records}.
\newblock \bibinfo{journal}{\emph{BMC medical informatics and decision making}}
  \bibinfo{volume}{8}, \bibinfo{number}{1} (\bibinfo{year}{2008}),
  \bibinfo{pages}{1--17}.
\newblock


\bibitem[\protect\citeauthoryear{of~Health and Services}{of~Health and
  Services}{[n.d.]}]%
        {hipaa2021}
\bibfield{author}{\bibinfo{person}{U.S.~Department of Health} {and}
  \bibinfo{person}{Human Services}.} \bibinfo{year}{[n.d.]}\natexlab{}.
\newblock \bibinfo{title}{U.S. Health Insurance Portability and Accountability
  Act (HIPAA) Privacy Rule.}
\newblock \bibinfo{howpublished}{\url{https://www.hhs.gov/hipaa/index.html}}.
\newblock
\newblock
\shownote{[Online; accessed 19-July-2021].}


\bibitem[\protect\citeauthoryear{Stubbs, Kotfila, and Uzuner}{Stubbs
  et~al\mbox{.}}{2015}]%
        {stubbs2015automated}
\bibfield{author}{\bibinfo{person}{Amber Stubbs}, \bibinfo{person}{Christopher
  Kotfila}, {and} \bibinfo{person}{{\"O}zlem Uzuner}.}
  \bibinfo{year}{2015}\natexlab{}.
\newblock \showarticletitle{Automated systems for the de-identification of
  longitudinal clinical narratives: Overview of 2014 i2b2/UTHealth shared task
  Track 1}.
\newblock \bibinfo{journal}{\emph{Journal of biomedical informatics}}
  \bibinfo{volume}{58} (\bibinfo{year}{2015}), \bibinfo{pages}{S11--S19}.
\newblock


\bibitem[\protect\citeauthoryear{Sweeney}{Sweeney}{1996}]%
        {sweeney1996replacing}
\bibfield{author}{\bibinfo{person}{Latanya Sweeney}.}
  \bibinfo{year}{1996}\natexlab{}.
\newblock \showarticletitle{Replacing personally-identifying information in
  medical records, the Scrub system.}. In \bibinfo{booktitle}{\emph{Proceedings
  of the AMIA annual fall symposium}}. American Medical Informatics
  Association, \bibinfo{pages}{333}.
\newblock


\bibitem[\protect\citeauthoryear{Toscano, O'Donnell, Unruh, Golinelli, Carullo,
  Messina, and Casalino}{Toscano et~al\mbox{.}}{2018}]%
        {toscano2018electronic}
\bibfield{author}{\bibinfo{person}{F Toscano}, \bibinfo{person}{E O'Donnell},
  \bibinfo{person}{MA Unruh}, \bibinfo{person}{D Golinelli}, \bibinfo{person}{G
  Carullo}, \bibinfo{person}{G Messina}, {and} \bibinfo{person}{LP Casalino}.}
  \bibinfo{year}{2018}\natexlab{}.
\newblock \showarticletitle{Electronic health records implementation: can the
  European Union learn from the United States?}
\newblock \bibinfo{journal}{\emph{European Journal of Public Health}}
  \bibinfo{volume}{28} (\bibinfo{year}{2018}), \bibinfo{pages}{213--401}.
\newblock


\bibitem[\protect\citeauthoryear{Tzeng, Hoffman, Saenko, and Darrell}{Tzeng
  et~al\mbox{.}}{2017}]%
        {tzeng2017adversarial}
\bibfield{author}{\bibinfo{person}{Eric Tzeng}, \bibinfo{person}{Judy Hoffman},
  \bibinfo{person}{Kate Saenko}, {and} \bibinfo{person}{Trevor Darrell}.}
  \bibinfo{year}{2017}\natexlab{}.
\newblock \showarticletitle{Adversarial discriminative domain adaptation}. In
  \bibinfo{booktitle}{\emph{Proceedings of the IEEE conference on computer
  vision and pattern recognition}}. \bibinfo{pages}{7167--7176}.
\newblock


\end{thebibliography}

\end{document}